\documentclass{article}
\date{}
\usepackage[english]{babel}

\usepackage[letterpaper,top=2cm,bottom=2.5cm,left=3cm,right=3cm,marginparwidth=1.75cm]{geometry}

\usepackage{setspace}
\usepackage{amssymb,amsmath,amsthm}
\usepackage{graphicx}
\usepackage[table,xcdraw]{xcolor}
\usepackage{multirow} 
\usepackage{caption}
\usepackage{subcaption}
\usepackage[colorlinks=true, allcolors=blue]{hyperref}
\usepackage{rotating}
\usepackage{adjustbox}
\usepackage[normalem]{ulem}
\useunder{\uline}{\ul}{}
\usepackage{lscape}
\usepackage{longtable}

\usepackage{hyperref}
\hypersetup{
colorlinks=true,
linkcolor=magenta,
citecolor=blue,
filecolor=magenta,
urlcolor=cyan,
}

\title{ArrhythmiaVision: Resource-Conscious Deep Learning Models with Visual Explanations for ECG Arrhythmia Classification}

\usepackage{authblk}
\author[1]{Zuraiz Baig}
\author[2]{Sidra Nasir}
\author[3]{Rizwan Ahmed Khan\thanks{Email corresponding author: rizwankhan@iba.edu.pk}}
\author[4]{Muhammad Zeeshan Ul Haque}

\affil[1]{Department of Electrical Engineering and Information Technology, Otto von Guerricke Universität, Magdeburg, Germany}
\affil[2]{Dipartimento di Informatica, Università di Verona, Verona, Italy}
\affil[3]{Department of Computer Science, School of Mathematics and Computer Science, Institute of Business Administration, Karachi, Pakistan}
\affil[4]{Faculty of Information Technology, Salim Habib University, Karachi, Pakistan}

\begin{document}
	\maketitle

	\begin{abstract} 
Cardiac arrhythmias are a leading cause of life-threatening cardiac events, highlighting the urgent need for accurate and timely detection. Electrocardiography (ECG) remains the clinical gold standard for arrhythmia diagnosis; however, manual interpretation is time-consuming, dependent on clinical expertise, and prone to human error. Although deep learning has advanced automated ECG analysis, many existing models abstract away the signal’s intrinsic temporal and morphological features, lack interpretability, and are computationally intensive—hindering their deployment on resource-constrained platforms. In this work, we propose two novel lightweight 1D convolutional neural networks, ArrhythmiNet V1 and V2, optimized for efficient, real-time arrhythmia classification on edge devices. Inspired by MobileNet’s depthwise separable convolutional design, these models maintain memory footprints of just 302.18 KB and 157.76 KB, respectively, while achieving classification accuracies of 0.99 (V1) and 0.98 (V2) on the MIT-BIH Arrhythmia Dataset across five classes: Normal Sinus Rhythm, Left Bundle Branch Block, Right Bundle Branch Block, Atrial Premature Contraction, and Premature Ventricular Contraction. The architectures are specifically designed to preserve the spatial morphology and temporal dynamics of ECG signals for robust performance. In order to ensure clinical transparency and relevance, we integrate Shapley Additive Explanations (SHAP) and Gradient-weighted Class Activation Mapping (Grad-CAM), enabling both local and global interpretability. These techniques highlight physiologically meaningful patterns—such as the QRS complex and T-wave—that contribute to the model’s predictions. We also discuss performance-efficiency trade-offs and address current limitations related to dataset diversity and generalizability. Overall, our findings demonstrate the feasibility of combining interpretability, predictive accuracy, and computational efficiency in practical, wearable, and embedded ECG monitoring systems. 
\end{abstract}



\section{Introduction} \label{sec:introduction}

Cardiovascular diseases (CVDs) remain the leading cause of death worldwide, accounting for millions of deaths each year \cite{worldhealthorganization_2019_cardiovascular}. Among these fatalities, sudden cardiac death (SCD) is particularly prevalent, often affecting individuals who appear relatively healthy and are under the age of 70 \cite{cnn_nearly}. Crucially, the risk of SCD diminishes significantly when arrhythmias electrophysiological anomalies in cardiac rhythm are detected and treated early \cite{buist_2002_effects}. However, diagnosing these arrhythmias is often time-consuming and requires specialized expertise, particularly when reliant on manual electrocardiogram (ECG) analysis. Consequently, the demand for automated and real-time arrhythmia detection systems has grown exponentially in both hospital settings and remote healthcare applications \cite{serhani_2020_ecg, sampson_2018_continuous, sajeev_2019_wearable}.

Over the past decade, advancements in machine learning (ML) and deep learning (DL) have significantly improved the accuracy of automated arrhythmia detection using ECG signals \cite{ebrahimi_2020_a}. Early studies employing traditional ML algorithms—such as Random Forests (RF), Support Vector Machines SVM), and Decision Trees (DT) showed promising performance gains but were often constrained by their reliance on hand-engineered features, which may fail to capture the intricate temporal and morphological patterns present in ECG signals. In contrast, DL models, particularly convolutional neural networks (CNNs) and recurrent neural networks (RNNs) learn robust feature representations directly from raw or minimally preprocessed signals, leading to higher accuracy and generalizability.

Despite these remarkable performance improvements, several challenges still impede the widespread clinical adoption of DL-based arrhythmia detection. One prominent hurdle is the “black box” nature of these models: they provide limited insight into their internal decision-making processes, which raises concerns about trust, accountability, and interpretability in the medical community \cite{madasu_2024_considerations}. To address this issue, researchers have begun exploring explainable artificial intelligence (XAI) methods that can render model predictions more transparent. While techniques like Class Activation Maps (CAM) \cite{zhou_2016_learning}, Gradient-weighted Class Activation Mapping (Grad-CAM) \cite{selvaraju_2020_gradcam}, Local Interpretable Model-agnostic Explanations (LIME) \cite{ribeiro_2016_why}, and Shapley Additive Explanations (SHAP) \cite{lundberg_2017_a} have been successfully employed in imaging or structured data settings \cite{chaddad_2023_survey}, comparatively fewer efforts have focused on applying these XAI strategies to one-dimensional biomedical signals such as ECG.

Another practical limitation is that many high-performing DL models are computationally heavy, posing challenges for deployment in real-time or resource-constrained environments such as wearable devices or embedded systems—where memory and processing power are limited. Reducing a model’s computational footprint without sacrificing diagnostic accuracy is therefore critical for next-generation ECG analysis tools.

This study proposes two novel lightweight one-dimensional CNN architectures, termed ArrhythmiNet V1 and V2, designed for arrhythmia classification from single-lead ECG signals. Both models draw inspiration from MobileNet architectures, known for their efficient depthwise separable convolutions that minimize parameter count and computational cost. To enhance clinical viability, we further integrate XAI techniques (SHAP and Grad-CAM), enabling a transparent understanding of the features and waveform regions influencing the models’ decisions. We demonstrate that these explainable, computationally lean models achieve state-of-the-art performance on the MIT-BIH arrhythmia dataset while maintaining a small memory footprint (302.18 KB and 157.76 KB for V1 and V2, respectively). These characteristics make ArrhythmiNet V1 and V2 well-suited for deployment in wearable cardiac monitors and other resource-constrained systems.

In the remainder of this paper, we first review related work in both ML and DL-based arrhythmia detection, along with existing efforts in XAI for biomedical signal analysis. Next, we describe the specifics of the proposed architectures and the dataset used for training and testing. We then discuss the integration of SHAP and Grad-CAM for model interpretability, followed by a detailed evaluation of classification performance and interpretability findings. Finally, we conclude by highlighting limitations and outlining avenues for future research in lightweight, explainable, and scalable ECG-based arrhythmia detection systems.

\section{Literature Review} \label{sec:literature_review}

Early ML techniques for ECG-based arrhythmia classification—such as decision trees, Naïve Bayes, and K-Nearest Neighbors (KNN) have generally struggled to exceed moderate performance thresholds. For instance, Soman and Bobbie \cite{soman_classification} reported accuracies below 80\% using OneR, a J48 decision tree, and Naïve Bayes for arrhythmia detection. Similarly, Saboori et al. \cite{saboori_2021_classification} experimented with multiple ML algorithms, including Naïve Bayes, Artificial Neural Networks (ANN), KNN, and SVM, but achieved maximum accuracies of only 71.86\% for K-NN and 70.70\% for SVM. These results exhibit the limitations of traditional ML approaches in capturing the complex morphological and temporal features inherent in ECG signals.

By contrast, deep learning (DL) has significantly improved classification performance due to its ability to automatically learn salient features directly from raw or minimally preprocessed signals. Hasan et al. \cite{hossain_2020_cardiac} demonstrated a 1D CNN achieving 98.28\% accuracy for five arrhythmias, while Al Rahhal et al. \cite{alrahhal_2018_convolutional} reported a 99.3\% accuracy using an LSTM architecture, although it was specialized for Atrial Fibrillation (AF). Other studies have employed various forms of RNNs, Long short term memory (LSTMs), and Gated Recurrent Units (GRUs) to detect AF, achieving accuracies up to 100\% in some cases \cite{sujadevi_2017_realtime, faust_2018_automated}. Wang et al. \cite{wang_2019_a} used a GRNN-LSTM model to attain a 99.8\% accuracy across four arrhythmias, whereas Rajkumar et al. \cite{rajkumar_2019_arrhythmia} employed a CNN that achieved 93.6\% accuracy over eight arrhythmia classes. Although these results highlight the power of DL architectures in ECG classification, they also raise concerns regarding their “black box” nature, which can impede clinical adoption when clinicians cannot readily interpret model outputs.

In this context, XAI methods have gained prominence for offering deeper insights into how DL models make decisions. In the domain of ECG classification, explainability approaches range from feature-based strategies to model-specific attribution maps. Some studies focus on feature-level reasoning. For example, Jo et al. \cite{jo_2021_explainable} emphasized rhythm irregularities and P-wave absence to interpret arrhythmias, while Sangroya et al. \cite{sangroya_2022_generating} generated clinician-friendly language explanations verified by cardiologists. Bender et al. \cite{bender_2023_analysis} employed first-order gradient-based relevance scores to delineate learned features across different beats and leads.

Alternatively, attribution and activation map techniques—such as Gradient-weighted Class Activation Mapping (Grad-CAM) and its variants are used to highlight localized signal regions crucial for classification. Gusarova et al. \cite{gusarova_explainable} and Varandas et al. \cite{varandas_2022_quantified} applied Grad-CAM-based approaches for myocardial infarction classification and general ECG anomaly detection, respectively. Other studies have compared multiple attribution techniques, including LIME, SHAP, and PSI, to identify robust explainability methods \cite{neves_2021_interpretable, zhang_interpretable, pawar_2020_incorporating}. Further adaptations include saliency map-based explanations for arrhythmia detection \cite{jo_2021_detection}, Grad-CAM for phonocardiogram analysis \cite{divakar_explainable}, and modified Grad-CAM in the LightX3ECG system to attain lead-wise interpretability \cite{le_2023_lightx3ecg}.

Despite these advances, two significant gaps remain. First, most studies either target a narrow subset of arrhythmias (e.g., atrial fibrillation or myocardial infarction) or do not rigorously evaluate multiple classes. Second, although high-performing deep models exist, relatively few prioritize lightweight architectures suitable for deployment on wearable or embedded systems. Consequently, there is a growing need for compact, high-accuracy models integrated with robust XAI methods to facilitate real-time, clinically interpretable ECG analysis in resource-constrained environments. 
\

\section{Methodology} \label{sec:Method}

This section presents the key components of the proposed framework, including the dataset and preprocessing pipeline (Section~\ref{sec:dataset}), the lightweight convolutional neural network (CNN) architectures (Section~\ref{sec:Arc}), and the explainable AI (XAI) techniques utilized (Section~\ref{sec:xai}). The methodology combines wavelet-based denoising, efficient depthwise separable convolutions inspired by MobileNet, and interpretability tools such as Grad-CAM and SHAP. Together, these components are designed to optimize the trade-off between predictive accuracy, computational efficiency, and clinical explainability.

\subsection{Dataset and Preprocessing}\label{sec:dataset}
\noindent Electrocardiogram (ECG) signal analysis is foundational to the detection and classification of cardiac arrhythmias. However, the reliability of machine learning models in this domain heavily depends on the quality, balance, and representation of the underlying data. In this study, we utilize the MIT-BIH Arrhythmia Database, a widely acknowledged benchmark for ECG-based arrhythmia classification. To ensure robust model performance, careful attention is given to preprocessing steps such as denoising, normalization, and class balancing. This section outlines the dataset characteristics, the challenges posed by its inherent class imbalance, and the signal processing techniques applied to prepare the data for model training. \newline
\begin{figure} [!htb]
   \centering
     \includegraphics[width=9cm, height=6cm]{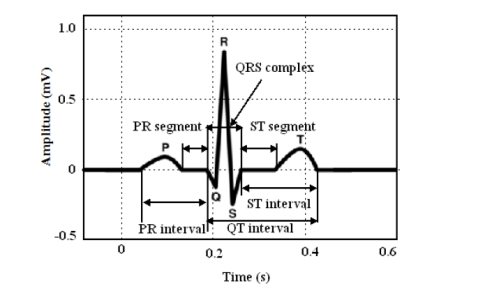}
     \caption{ The standard PQRST waveform of an electrocardiogram (ECG)}
     \label{fig: ECG}
\end{figure}

{\it{The ECG Signal:}} The electrocardiogram (ECG) is a 
vital diagnostic tool that records the electrical activity of the heart. It consists of key waveform components, each representing a specific physiological event. The P wave corresponds to atrial depolarization, initiating atrial contraction. The PR interval reflects the time taken for electrical impulses to travel from the atria to the ventricles. The QRS complex represents rapid ventricular depolarization, leading to ventricular contraction, while the ST segment denotes the interval between depolarization and repolarization. The T wave signifies ventricular repolarization, restoring the resting state of the heart. These waveform landmarks are represented by chen et al. \cite{Yol_2019_Design} in figure \ref{fig: ECG}. 

Alterations in ECG waveform morphology or timing can indicate cardiac abnormalities such as arrhythmias, ischemia, or electrolyte imbalances. A prolonged QT interval is associated with an increased risk of life-threatening arrhythmias. The ECG's ability to provide real-time insights into cardiac function makes it an indispensable tool for diagnosing and monitoring heart conditions. Understanding the physiological significance of each ECG component aids in accurate interpretation and effective clinical decision-making.

\begin{figure}[!htb]
   \centering
   \includegraphics[width=0.65\textwidth]{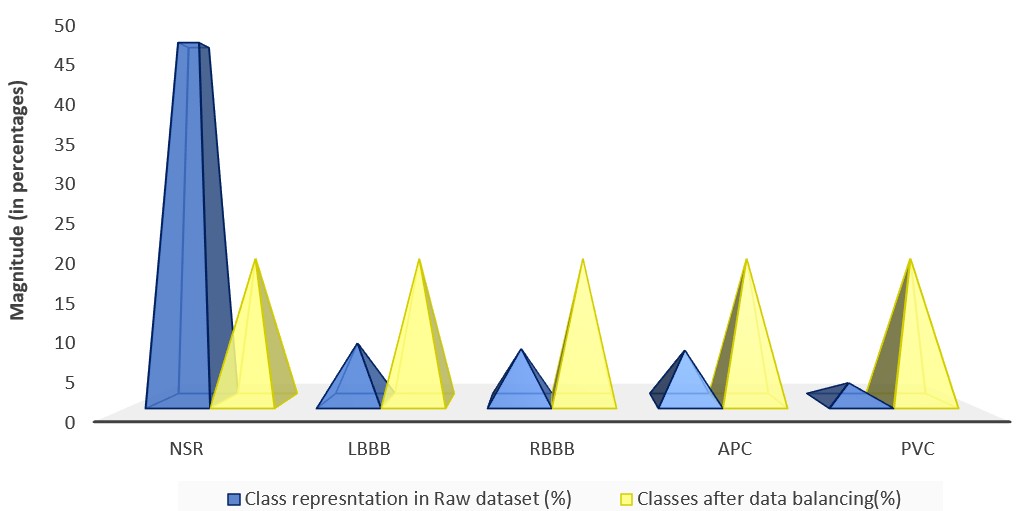}
   \caption{(Left) Distribution of the unbalanced raw MIT-BIH dataset. (Right) Distribution after applying balancing techniques.}
   \label{fig: datadistribution}
\end{figure}

\subsubsection{MIT-BIH Arrhythmia Dataset}
\begin{figure}[!htb] 
   \centering
     \includegraphics[width=0.75\textwidth]{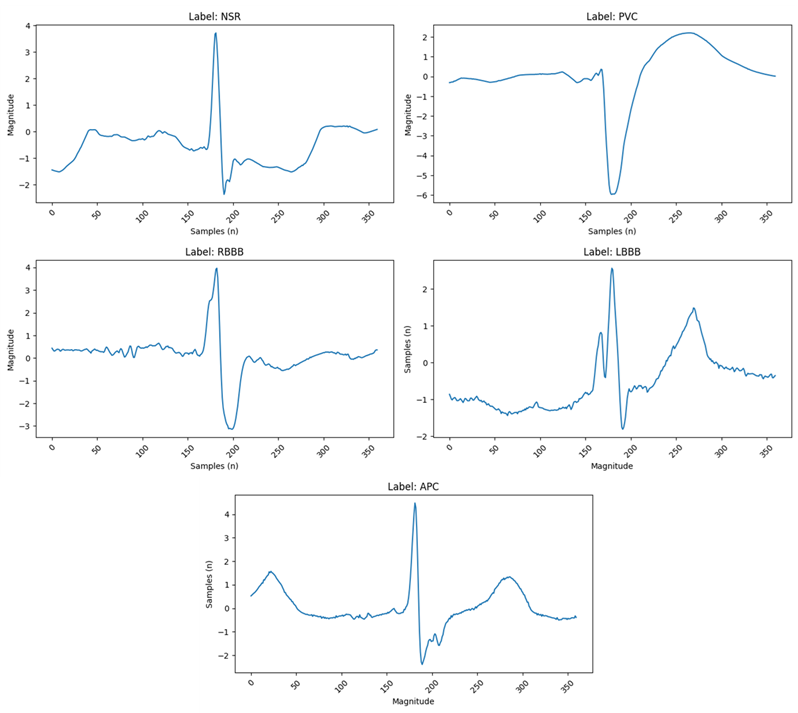}
     \caption{(Top Left) Normal Sinus Rhythm, the standard ECG signal exhibiting all regular morphological features. (Top Right) Premature Ventricular Contraction, characterized by an early contraction of the ventricles. (Middle Left) Right Bundle Branch Block, occurring due to a blockage in the right bundle branch, preventing proper action potential propagation. (Middle Right) Left Bundle Branch Block, resulting from a similar blockage but affecting the left conduction pathway. (Bottom) Atrial Premature Contraction, an early contraction of the atrial chambers of the heart. }
     \label{fig: Data}
\end{figure}

The publicly available MIT-BIH Arrhythmia Database \cite{goldberger_physiobank} is utilized in this study as it is widely recognized for cardiac arrhythmia research. It contains approximately 1.1 million beats sampled at 360 Hz, each labelled by multiple cardiologists. The five target classes are \textit{Normal Sinus Rhythm} (NSR), \textit{Left-Bundle Branch Block} (LBBB), \textit{Right Bundle Branch Block} (RBBB), \textit{Atrial Premature Contraction} (APC), and  \textit{Premature Ventricular Contraction} (PVC), the morphology and samples of which are shown in fig \ref{fig: Data}.

Despite its popularity, the dataset exhibits significant class imbalance, as illustrated in Fig.~\ref{fig: datadistribution}. To mitigate bias, oversampling of minority classes and undersampling of the majority classes were performed before model training. After balancing, the data were split into training (80\%) and testing (20\%) subsets, ensuring that each arrhythmia class was adequately represented.

\subsubsection{Wavelet-Based Denoising and Normalization}
Before splitting the dataset, each ECG signal undergoes denoising using a wavelet transform with Symlet 4. Specifically, the signal is decomposed into approximation and detail coefficients, and a threshold is applied to the detail coefficients to suppress noise while retaining clinically relevant morphology. The denoised signal is then reconstructed and normalized, which stabilizes training by maintaining consistent amplitude ranges across different beats and subjects. Notably, this wavelet-based approach helps preserve characteristic features such as QRS complexes and T-wave segments that are crucial for accurate arrhythmia detection \cite{ang_2023_a, gao_2019_an, singh_2018_sciencedirectncnd, raza_2022_designing}.

\subsection{1D CNN Architecture}\label{sec:Arc}

Two custom 1D CNN architectures—ArrhythmiNet V1 and ArrhythmiNet V2—were designed to achieve high diagnostic performance with minimal computational overhead. Both draw inspiration from MobileNet \cite{howard_mobilenets, sandler_mobilenetv2}, employing depthwise separable convolutions to reduce the parameter count. Unlike the 2D kernels in traditional MobileNet, these architectures use 1D convolutions optimized for time-series ECG signals.

\subsubsection{ArrhythmiNet V1}  

\begin{figure}[!htb]
   \centering
   \includegraphics[width=15cm,height=4cm]{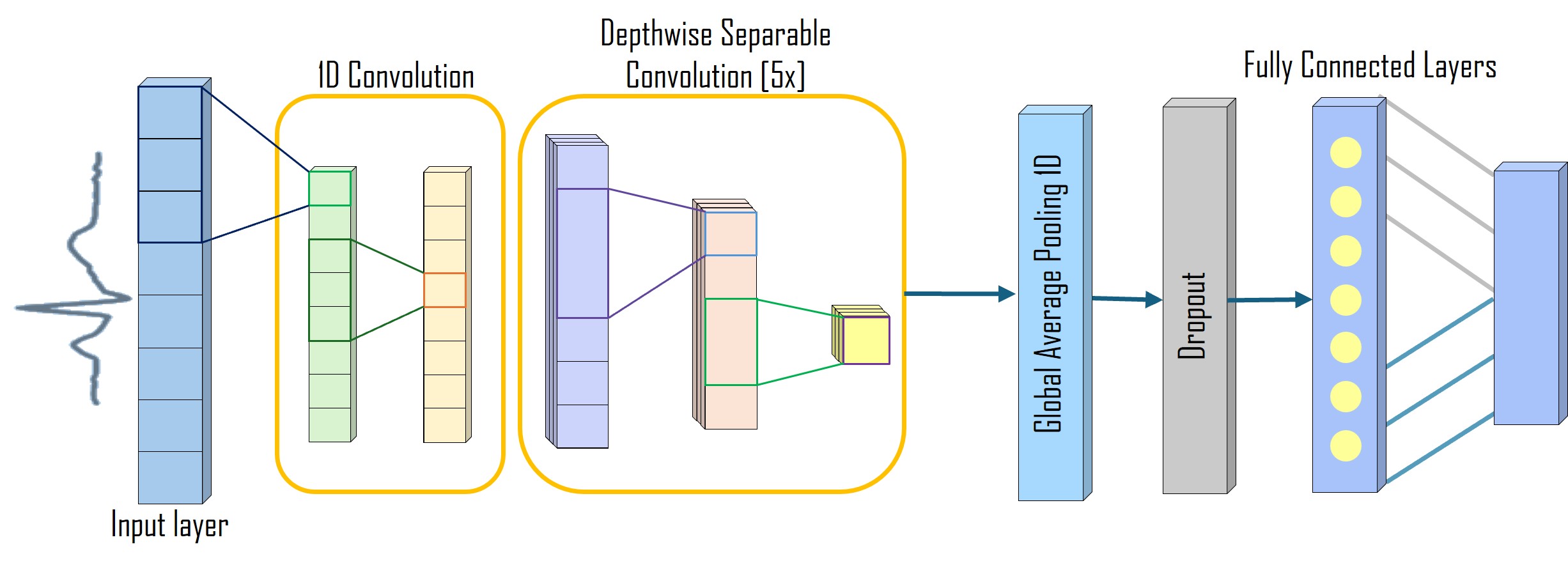}
   \caption{ArrhythmiNet V1 architecture featuring five depthwise separable convolution blocks.}
   \label{fig: arch1}
   \end{figure}
   
ArrhythmiNet V1 addresses the need for a compact yet effective network for ECG analysis. As shown in Fig.~\ref{fig: arch1}, the model begins with an input layer optimized for ECG waveforms (sampled at 360 Hz) and processes the signal through five sequential convolutional blocks. Each block contains a standard 1D convolutional layer (with kernel size \(k\)), followed by batch normalization and ReLU activation, and then a depthwise separable convolution block—comprising a depthwise convolution (with \(k \times C_{\mathrm{in}}\) parameters) followed by a pointwise 1D convolution (with \(C_{\mathrm{in}} \times C_{\mathrm{out}}\) parameters) along with their associated batch normalization and ReLU activations. This two-stage operation substantially reduces the parameter count compared to a conventional 1D convolution, which would require parameters expressed in the following equation:

\begin{equation}
\frac{P_{\mathrm{DS}}}{P_{\mathrm{standard}}} = \frac{1}{C_{\mathrm{out}}} + \frac{1}{k},
\end{equation}

where \(P_{\mathrm{DS}}\) is the parameter count for the depthwise separable convolution. This formulation highlights that the depthwise separable approach dramatically reduces the number of parameters—especially for larger values of \(k\) and \(C_{\mathrm{out}}\). Finally, a global average pooling layer condenses the spatiotemporal features before a softmax-based classification is applied. Experimental results show that ArrhythmiNet V1 occupies only 303~KB in its uncompressed form, emphasizing the efficiency gains achieved through the use of depthwise separable convolutions.

\subsubsection{ArrhythmiNet V2}
\begin{figure}[!ht]
   \centering
   \includegraphics[width=16cm,height=4cm]{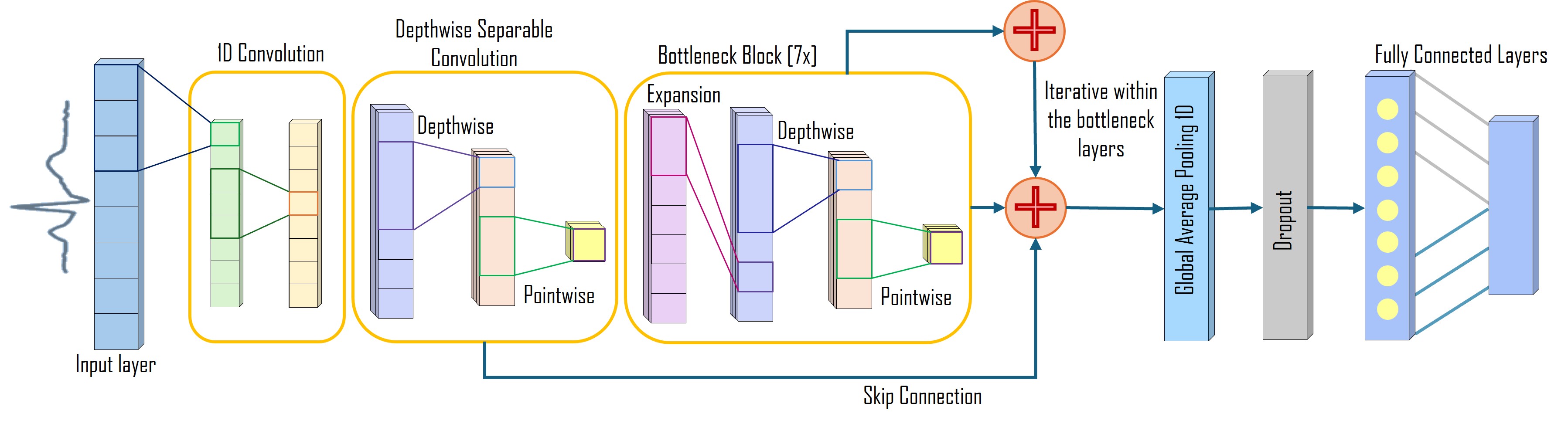}
   \caption{ArrhythmiNet V2 architecture showing residual connections and seven bottleneck blocks.}
   \label{fig: arch2}
\end{figure}

Building on V1, ArrhythmiNet V2 incorporates bottleneck blocks and skip connections to further enhance gradient flow and feature extraction, as depicted in Fig.~\ref{fig: arch2}. Each bottleneck block comprises of pointwise (1D) expansion layer that increases the number of channels, a depthwise separable convolution for channel-specific filtering, and a pointwise projection layer that reduces the dimensionality back to a smaller channel space.

\begin{table}[!ht]
\centering
\caption{Training Duration and Model Size}
\begin{tabular}{|c|c|c|c|c|}
\hline
\textbf{Model} & \textbf{Epochs} & \textbf{Mini-Batches} & \textbf{Cumulative (s)} & \textbf{Time/Epoch (s)} \\ \hline
ArrhythmiNet V1 & 30 & 500 & $\sim$2576 & $\sim$86 \\
ArrhythmiNet V2 & 30 & 500 & $\sim$1767 & $\sim$59 \\ \hline
\end{tabular}
\label{Tab : compdet}
\end{table}

\textbf{Utility of Depthwise Convolutions}
Depthwise separable convolutions reduce parameter counts by separating spatial filtering from channel mixing. For 1D ECG signals, a depth-wise convolution treats each input channel independently, followed by a point-wise convolution that fuses channel information. As summarized in Table~\ref{Tab:compconv}, this significantly reduces the multiply accumulate (MAC) operations relative to standard convolutions. Over multiple layers, the reduction becomes pronounced, making the ArrhythmiNet architectures well suited for embedded or wearable applications that require real-time inference.

\begin{table}[!ht]
\centering
\caption{Comparison of Standard and Depthwise Convolution Costs (1D Case)}
\begin{tabular}{|c|c|c|}
\hline
\textbf{Convolution Type} & \textbf{Parameters} & \textbf{MAC Operations} \\ \hline
Standard 1D & $k \times C_{\mathrm{in}} \times C_{\mathrm{out}}$ & $L' \times C_{\mathrm{out}} \times k \times C_{\mathrm{in}}$ \\
Depthwise 1D & $k \times C_{\mathrm{in}}$ & $L' \times C_{\mathrm{in}} \times k$ \\ \hline
\end{tabular}
\label{Tab:compconv}
\end{table}

\textbf{Skip Connections}
The ArrhythmiNet V2 includes skip connections to combat feature degradation in deeper networks. A typical transformation can be expressed as 
\begin{equation}
Y^{(c)} = f\bigl(W^{(c)} * X^{(c)} + b^{(c)}\bigr) + X^{(c)},
\end{equation}
where $X^{(c)}$ is the input signal per channel $c$, $W^{(c)}$ is the depthwise filter, and $f(\cdot)$ is the activation function. These connections ensure that gradient signals remain strong during backpropagation, preserving subtle ECG waveforms critical for clinical differentiation between arrhythmias.

The residual connections mitigate vanishing gradients and help preserve morphological details critical for distinguishing arrhythmias such as LBBB, RBBB, and PVC. Despite adding these connections, V2 occupies only 158KB almost half of V1. Table~\ref{Tab : compdet} outlines the training details in a system equipped with dual Intel Xeon CPUs (2.20 GHz, hyperthreading enabled). Notably, ArrhythmiNet V2 converged faster (59 seconds/epoch) compared to V1 (86 seconds/epoch), offering a competitive blend of speed, parameter efficiency, and accuracy.

\subsection{Explainable AI (XAI) Techniques} \label{sec:xai}

Reliable medical AI solutions should complement high performance with transparent decision-making processes. Consequently, two complementary XAI approaches—Grad-CAM and SHAP—were chosen for post-hoc interpretability.

\subsubsection{Grad-CAM for 1D Signals}
Gradient-weighted Class Activation Mapping (Grad-CAM) \cite{selvaraju_2020_gradcam} is extended to 1D ECG. We compute the gradient of the predicted class score $y^c$ w.r.t. the final convolutional feature maps $A^k$, global-average-pool these gradients to obtain $\alpha_k^c$, and apply:
\begin{equation}
L_{\mathrm{Grad-CAM}}^c = \mathrm{ReLU}\Bigl(\sum_k \alpha_k^c \cdot A^k \Bigr).
\end{equation}
This produces a heatmap aligned to the ECG timeline, highlighting intervals (e.g., QRS complex, T-wave region) that most influenced the model’s classification.

\subsubsection{SHAP for Feature Attribution}
SHAP (SHapley Additive exPlanations) \cite{lundberg_2017_a} calculates the contribution of each feature (time step in the ECG) to a model’s prediction via cooperative game theory. Formally, for a feature $i$:
\begin{equation}
\phi_i = \sum_{S \subseteq N \setminus \{i\}} \frac{|S|!\,(|N| - |S| - 1)!}{|N|!} \bigl[f(S \cup \{i\}) - f(S)\bigr],
\end{equation}
Where $N$ is the set of all features, these Shapley values can be aggregated to show how strongly each segment of the ECG waveform contributes to a classification decision, offering a local and global interpretative perspective.

The integration of Grad-CAM and SHAP yields a multifaceted view of model decisions. Grad-CAM’s visually interpretable maps expose the high-level temporal segments deemed important, whereas SHAP delves deeper, assigning quantitative contributions to individual time points. Consistency in these attributions across methods bolsters confidence in the model’s reliability and clinical plausibility, while discrepancies can guide further debugging or model refinement. This dual-layered analysis is crucial for AI in healthcare, where transparent and justifiable decisions are prerequisites for clinical acceptance and integration.

\section{Results}

The experimental evaluation was conducted on the MIT-BIH Arrhythmia Dataset, to assess both the predictive performance and interpretability of the proposed ArrhythmiNet architectures. In this section, we present quantitative metrics such as precision, recall, and F1-score for each arrhythmia class, along with confusion matrices that detail the distribution of classification errors. Additionally, we provide a qualitative analysis of the model explanations using state-of-the-art XAI methods (SHAP and Grad-CAM). These results not only demonstrate the high accuracy and efficiency of the proposed models but also emphasis on their clinical relevance by highlighting key morphological features in the ECG signals that inform the model’s decisions. The following subsections detail the model performance and the interpretability assessments.

\subsection{Model Performance}

The models in this study were trained with adjusted hyperparameters to achieve the best performance for the medical classification task. Given the clinical nature of the problem, evaluation metrics such as recall, precision, and F1 score were prioritized to minimize false negatives and ensure reliable results.

As shown in Tables \ref{Tab: classificationreportV1} and \ref{Tab: classificationreportV2}, the ArrhythmiNet V1 model slightly outperformed the ArrhythmiNet V2 architecture despite its residual connections. This was reflected in higher average accuracy and better per-class recall, highlighting the strengths of convolutional neural networks (CNNs) in identifying relevant features. A close look at the Evaluation metrics for both the models highlights the fact the models have almost comparable performance for all three metrics when it is about only the arrhythmia classes i.e. LBBB, RBBB, APC and PVC. The main difference lies in the performance with the detection of NSR for ArrhythmiNet V2, this implies that the model is more prone and sensitive to the diseased classes of signals thus allowing the room for further investigation and care to be conducted by the clinicians. 

\begin{figure} 
   \centering
     \includegraphics[width=1.0\textwidth]{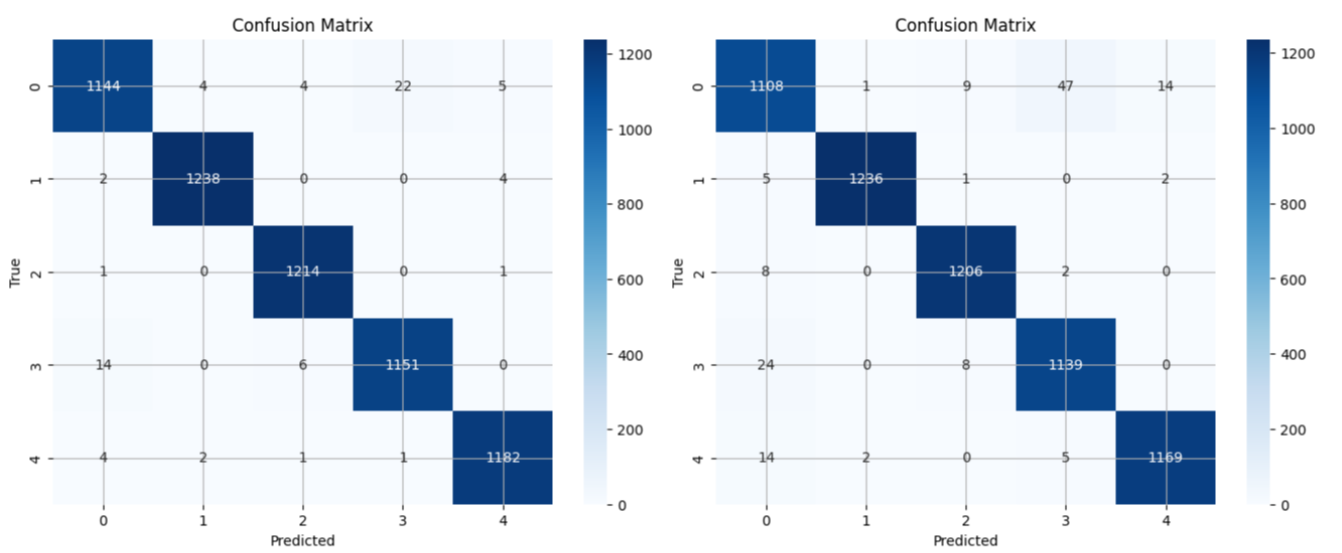}
     \caption{ (Left) Confusion matrix for ArrhythmiNet V1, showing minimal misclassifications across all classes. (Right) Confusion matrix for ArrhythmiNet V2, exhibiting slightly lower accuracy than the former but still yielding promising results.}
     \label{fig: confusionmatrix}
\end{figure}

The confusion matrices in Figure \ref{fig: confusionmatrix} offer a clearer picture of how both models performed in classifying different heart rhythms. One of the key challenges observed was distinguishing between NSR and LBBB. The difficulty in differentiating these rhythms suggests that they share similar morphological features, making it harder for the models to separate them accurately.  

The comparative analysis of the false negatives and false positives for each class outlined in Table \ref{Tab : false classifications} show that indeed the model is quite accurate it signifies that only 35 diseased instances were falsely classified as NSR, while 51 instances of NSR were falsely classified as Arrhythmic, this signifies the model's affinity to the arrhythmic classes.  Between the two models, ArrhythmiNet V2 had slightly more difficulty in correctly identifying these rhythms compared to ArrhythmiNet V1. This indicates that ArrhythmiNet V1 may have been better at capturing subtle differences in ECG waveforms, allowing for more accurate classification. Despite this challenge, both models performed well in recognizing distinct waveform characteristics, which played a crucial role in their overall classification accuracy.  

These findings highlight that while both models were effective in identifying unique signal patterns, certain rhythm types remained more challenging to classify. Refinements in feature extraction and preprocessing techniques could improve their ability to differentiate closely related cardiac conditions. Additionally, these results highlight that morphological features play a larger role in influencing model performance than temporal dependencies. CNN-based models effectively leverage these visual features, underlining the importance of choosing architectures suited to the data type and task requirements. These results demonstrate the potential for achieving state-of-the-art classification performance with a significantly reduced parameter thus promoting integration of deep learning in wearables and resource-constrained computational devices.

\begin{table}[ht]
\centering
\begin{minipage}{0.45\linewidth}
\centering
\caption{Classification Report Mobile Net V1}
\label{tab:mobv1}
\begin{tabular}{c c c c c}
\hline
\textbf{Class} & \textbf{Precision} & \textbf{Recall} & \textbf{F1-Score} & \textbf{Support} \\
\hline
N & 0.98 & 0.97 & 0.98 & 1179 \\
L & 1.00 & 1.00 & 1.00 & 1244 \\
R & 0.99 & 1.00 & 0.99 & 1216 \\
A & 0.98 & 0.98 & 0.98 & 1171 \\
V & 0.99 & 0.99 & 0.99 & 1190 \\
\hline
\multicolumn{4}{l}{\textbf{Accuracy}} & 0.99 \\
\multicolumn{4}{l}{\textbf{Macro Avg}} & 0.99 \\
\multicolumn{4}{l}{\textbf{Weighted Avg}} & 0.99 \\
\hline
\multicolumn{5}{l}{\textbf{Total Support: 6000}} \\
\hline
\label{Tab: classificationreportV1}
\end{tabular}
\end{minipage}
\hfill
\begin{minipage}{0.45\linewidth}
\centering
\caption{Classification Report Mobile Net V2}
\label{tab:mobv2}
\begin{tabular}{c c c c c}
\hline
\textbf{Class} & \textbf{Precision} & \textbf{Recall} & \textbf{F1-Score} & \textbf{Support} \\
\hline
N & 0.96 & 0.94 & 0.95 & 1179 \\
L & 1.00 & 0.99 & 1.00 & 1244 \\
R & 0.99 & 0.99 & 0.99 & 1216 \\
A & 0.95 & 0.97 & 0.96 & 1171 \\
V & 0.99 & 0.98 & 0.98 & 1190 \\
\hline
\multicolumn{4}{l}{\textbf{Accuracy}} & 0.98 \\
\multicolumn{4}{l}{\textbf{Macro Avg}} & 0.98 \\
\multicolumn{4}{l}{\textbf{Weighted Avg}} & 0.98 \\
\hline
\multicolumn{5}{l}{\textbf{Total Support: 6000}} \\
\hline
\label{Tab: classificationreportV2}
\end{tabular}
\end{minipage}
\end{table}

\begin{table}[ht]
\centering
\caption{Analysis of False Classifications}
\label{tab:false_classifications}
\begin{tabular}{p{0.8cm} p{1.2cm} p{0.5cm} p{0.5cm} p{5cm} c c p{5cm}}
\hline
\textbf{Class} & 
\textbf{Support} & 
\textbf{FP} & 
\textbf{FP (\%)} & 
\textbf{FP breakdown} & 
\textbf{FN} & 
\textbf{FN (\%)} & 
\textbf{FN breakdown} \\
\hline \\
NSR  & 1179 & 35 & 0.58 & \{'LBBB': 7, 'RBBB': 3, 'APC': 16, 'PVC': 9\} & 51 & 4.33 & \{'LBBB': 6, 'RBBB': 2, 'APC': 28, 'PVC': 15\} \\
LBBB & 1244 & 9  & 0.15 & \{'NSR': 6, 'PVC': 3\}                        & 13 & 1.05 & \{'NSR': 7, 'APC': 1, 'PVC': 5\}                       \\
RBBB & 1216 & 3  & 0.05 & \{'NSR': 2, 'APC': 1\}                          & 13 & 1.07 & \{'NSR': 3, 'APC': 6, 'PVC': 4\}                       \\
APC  & 1171 & 36 & 0.60 & \{'NSR': 28, 'LBBB': 1, 'RBBB': 6, 'PVC': 1\}   & 17 & 1.45 & \{'NSR': 16, 'RBBB': 1\}                               \\
PVC  & 1190 & 24 & 0.40 & \{'NSR': 15, 'LBBB': 5, 'RBBB': 4\}             & 13 & 1.09 & \{'NSR': 9, 'LBBB': 3, 'APC': 1\}                       \\
\hline
\label{Tab : false classifications}

\end{tabular}
\end{table}

\subsection{XAI Implementation}

The methods for ensuring the explainability of the models were SHAP and Grad-CAM. These methods are some of the most renowned and intuitive methods for developing an understanding of the model's internal workings. 

\begin{figure} 
   \centering
     \includegraphics[width=1.0\textwidth]{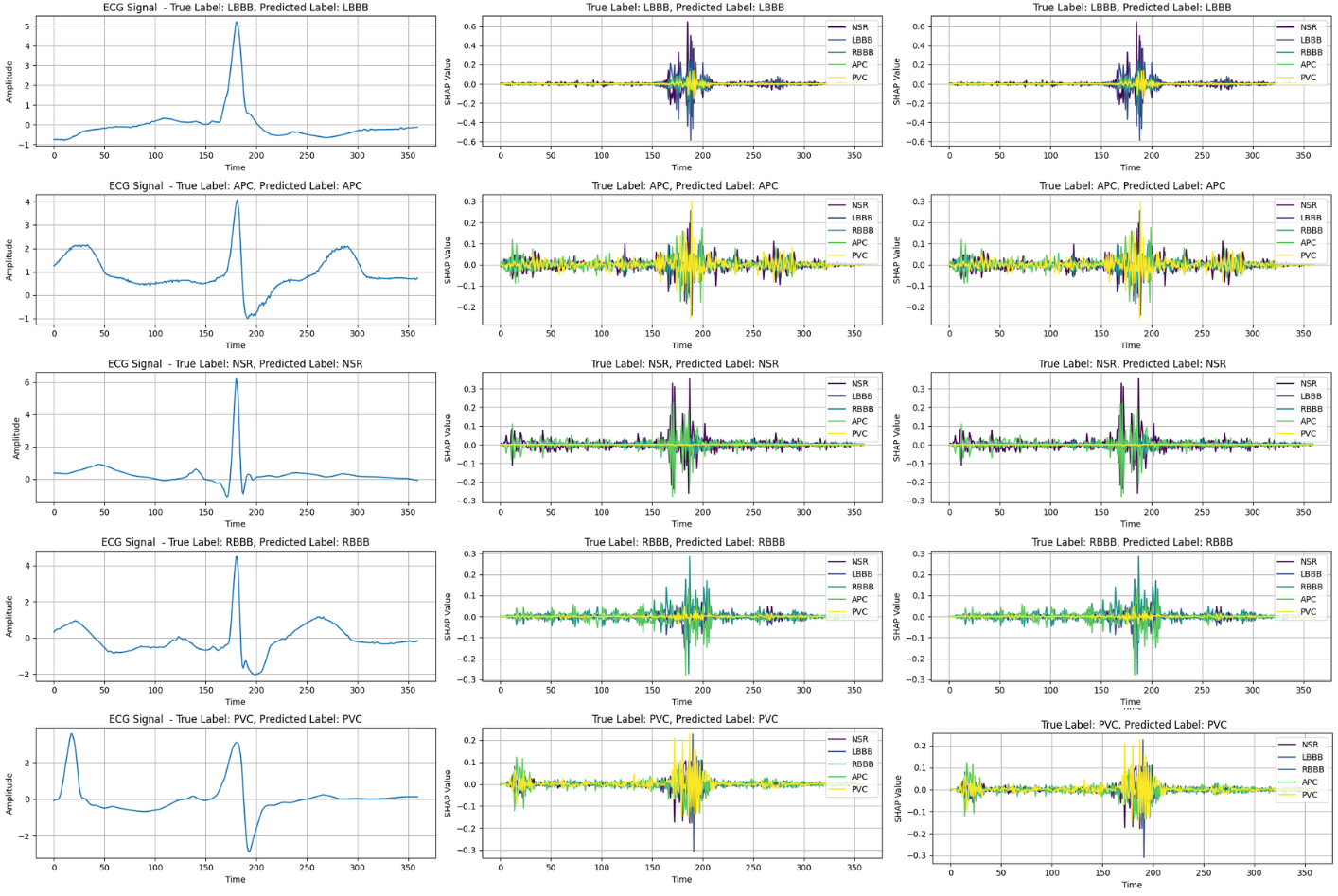}
     \caption{ ((Left) Randomly sampled original ECG signal. (Middle) SHAP output for ArrhythmiNet V1. (Right) SHAP output for ArrhythmiNet V2.}
     \label{fig: SHAP}
\end{figure}

A comparative analysis of the SHAP outputs for ArrhythmiNet V2 and ArrhythmiNet V1 reveals distinct operational characteristics and diagnostic efficacy in the context of ECG classification as seen in Figure \ref{fig: SHAP}. It's pertinent to mention here, that certain characteristics of ECG are used by clinicians to distinguish ECG Arrhythmias and the analysis of the models has been done qualitatively through the usage of the same \cite{docherty_2003_12lead} \cite{auricchio_2004_characterization} \cite{tzogias_2014_electrocardiographic}. Both models demonstrate competence in identifying critical regions of the ECG signal particularly the QRS complex and P wave that are essential for distinguishing arrhythmias by established cardiological criteria. ArrhythmiNet V2 exhibits robust feature attribution, with pronounced SHAP peaks corresponding to classical morphological markers such as the “M” pattern in LBBB, the “RSR” configuration in RBBB, and the signature wide QRS with an absent P wave in PVC. Nevertheless, its SHAP responses become more diffuse in cases of misclassification, suggesting a challenge in resolving subtle differences between overlapping waveform features.

In contrast, ArrhythmiNet V1 presents more localized SHAP activations, concentrating its attributions on narrower segments of the ECG waveform. This focused approach may enhance precision in identifying features like the early P wave in APC; however, it potentially limits the model’s ability to capture the broader context necessary for the accurate interpretation of complex arrhythmias, as evidenced by less pronounced responses in PVC cases. These findings indicate that while ArrhythmiNet V2 benefits from a comprehensive feature focus, it may require further refinement to improve class discrimination among morphologically similar arrhythmias. Conversely, ArrhythmiNet V1’s precision in feature localization, though beneficial for certain arrhythmia types, might be augmented by incorporating a more holistic interpretative strategy.

\begin{figure} 
   \centering
     \includegraphics[width=1.0\textwidth]{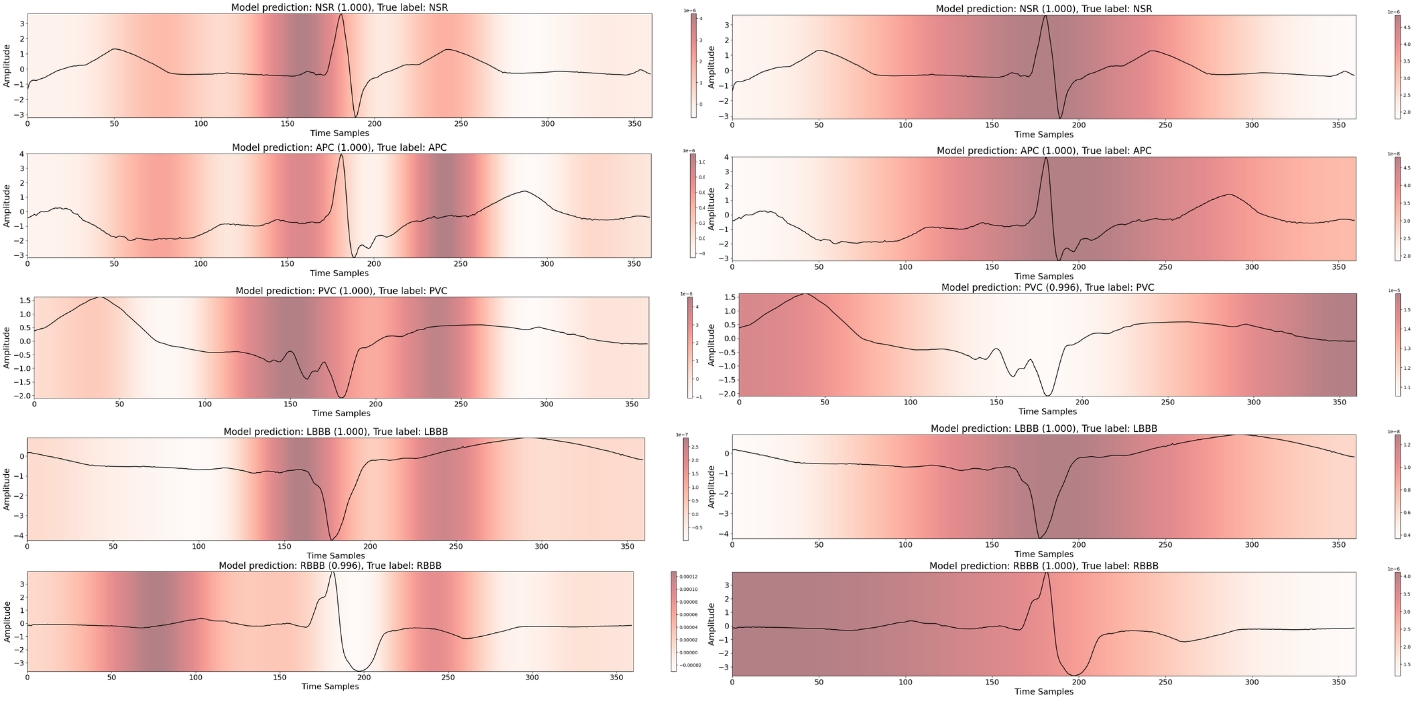}
    \caption{ (Left) Grad-CAM output for ArrhythmiNet V1. (Right) Grad-CAM results for ArrhythmiNet V2.}
    \label{fig: GradCAM}
\end{figure}

Similarly, From a critical clinical perspective, the Grad-CAM analysis of ArrhythmiNet V1 and ArrhythmiNet V2, seen in Figure \ref{fig: GradCAM} offer complementary insights into their diagnostic strategies in ECG classification. ArrhythmiNet V1’s architecture with a last convolutional output of (1, 10, 120) and an upsampled heatmap of 360-time points enables the model to capture highly localized features. This precise focus facilitates the identification of distinct arrhythmic patterns such as the “M” configuration in LBBB and the “RSR” pattern in RBBB, which are essential for clinical differentiation. However, the constrained temporal window has limited the detection of more diffuse morphological alterations observed in conditions such as "PVC", where a broader waveform context is critical for accurate diagnosis.

ArrhythmiNet V2’s design, characterized by a last convolutional output of (1, 3, 190) and a similarly upsampled 360-point heatmap, emphasizes extended temporal coverage at the expense of channel depth. This broader temporal focus affords a more comprehensive assessment of ECG signals, thus, facilitating the capture of subtle, distributed features inherent in arrhythmias such as "APC" and "PVC". While, this holistic approach improves the model’s ability to consider the full morphological evolution of a beat, it may also reduce the precision with which class-specific attributes such as the localized “M” pattern in LBBB—are delineated.

In light of established clinical criteria for arrhythmia interpretation, the differences in Grad-CAM activations are significant to the trade-off between localization and temporal comprehensiveness. ArrhythmiNet V1’s targeted feature extraction aligns well with conditions where distinct, localized abnormalities are diagnostically critical, whereas ArrhythmiNet V2’s extensive temporal perspective may offer advantages in scenarios requiring the integration of subtle and distributed signal variations. These findings suggest that an optimal ECG classification system might benefit from a hybrid strategy that synthesizes the strengths of both approaches, thereby enhancing diagnostic accuracy and clinical relevance in arrhythmia detection.

\section{Conclusion}

In this study, we introduced two novel lightweight 1D CNN architectures, ArrhythmiNet V1 and V2, for efficient ECG arrhythmia classification. By incorporating depthwise separable convolutions and residual connections drawing inspiration from MobileNet our models achieve state-of-the-art classification accuracy while maintaining exceptionally low memory footprints (303.18 KB for V1 and 157.76 KB for V2). The carefully designed preprocessing pipeline, which employs wavelet-based denoising and signal normalization, proved critical for preserving the ECG's morphological features that are essential for accurate diagnosis. Furthermore, the integration of explainable AI techniques, specifically Grad-CAM and SHAP, provided a detailed insight into the models' decision-making processes, ensuring that the automated predictions are consistent with established clinical ECG characteristics.

The current work is limited to single-lead ECG signals, which may not fully capture the diagnostic information available in multi-lead recordings. Future work will address this limitation by validating our models on external datasets, refining the XAI framework with more rigorous quantitative assessments, and extending the architectures to process multi-lead ECG inputs. In addition, exploring hybrid strategies that amalgamate the strengths of both ArrhythmiNet V1 and V2 could further enhance overall performance and interpretability. Such developments are anticipated to facilitate the deployment of deep learning-based arrhythmia detection systems in wearable and embedded devices, thereby enabling real-time, clinically interpretable cardiac monitoring in resource-constrained environments.

\addcontentsline{toc}{chapter}{References}
\bibliographystyle{ieeetr} 

\begin{thebibliography}{10}
	
	\bibitem{worldhealthorganization_2019_cardiovascular}
	W.~H. Organization, ``Cardiovascular diseases,'' 06 2019.
	
	\bibitem{cnn_nearly}
	S.~S. CNN, ``Nearly half of us adults have cardiovascular disease, study
	says.''
	
	\bibitem{buist_2002_effects}
	M.~D. Buist, ``Effects of a medical emergency team on reduction of incidence of
	and mortality from unexpected cardiac arrests in hospital: preliminary
	study,'' {\em BMJ}, vol.~324, pp.~387--390, 02 2002.
	
	\bibitem{serhani_2020_ecg}
	M.~A. Serhani, H.~T.~El~Kassabi, H.~Ismail, and A.~Nujum~Navaz, ``Ecg
	monitoring systems: Review, architecture, processes, and key challenges,''
	{\em Sensors}, vol.~20, p.~1796, 03 2020.
	
	\bibitem{sampson_2018_continuous}
	M.~Sampson, ``Continuous ecg monitoring in hospital: part 2, practical
	issues,'' {\em British Journal of Cardiac Nursing}, vol.~13, pp.~128--134, 03
	2018.
	
	\bibitem{sajeev_2019_wearable}
	J.~K. Sajeev, A.~N. Koshy, and A.~W. Teh, ``Wearable devices for cardiac
	arrhythmia detection: a new contender?,'' {\em Internal Medicine Journal},
	vol.~49, pp.~570--573, 05 2019.
	
	\bibitem{ebrahimi_2020_a}
	Z.~Ebrahimi, M.~Loni, M.~Daneshtalab, and A.~Gharehbaghi, ``A review on deep
	learning methods for ecg arrhythmia classification,'' {\em Expert Systems
		with Applications: X}, vol.~7, p.~33, 2020.
	
	\bibitem{madasu_2024_considerations}
	S.~Madasu and R.~Madasu, ``Considerations and challenges in deep learning and
	ai adoption in healthcare considerations and challenges in deep learning and
	ai adoption in healthcare,'' 2024.
	
	\bibitem{zhou_2016_learning}
	B.~Zhou, A.~Khosla, A.~Lapedriza, A.~Oliva, and A.~Torralba, ``Learning deep
	features for discriminative localization,'' 2016.
	
	\bibitem{selvaraju_2020_gradcam}
	R.~R. Selvaraju, M.~Cogswell, A.~Das, R.~Vedantam, D.~Parikh, and D.~Batra,
	``Grad-cam: Visual explanations from deep networks via gradient-based
	localization,'' {\em International Journal of Computer Vision}, vol.~128,
	p.~336–359, 02 2020.
	
	\bibitem{ribeiro_2016_why}
	M.~T. Ribeiro, S.~Singh, and C.~Guestrin, ``"why should i trust you?",'' {\em
		Proceedings of the 22nd ACM SIGKDD International Conference on Knowledge
		Discovery and Data Mining - KDD '16}, pp.~1135--1144, 2016.
	
	\bibitem{lundberg_2017_a}
	S.~Lundberg, P.~Allen, and S.-I. Lee, ``A unified approach to interpreting
	model predictions,'' 2017.
	
	\bibitem{chaddad_2023_survey}
	A.~Chaddad, J.~Peng, J.~Xu, and A.~Bouridane, ``Survey of explainable ai
	techniques in healthcare,'' {\em Sensors}, vol.~23, p.~634, 01 2023.
	
	\bibitem{soman_classification}
	T.~Soman and P.~Bobbie, ``Classification of arrhythmia using machine learning
	techniques.''
	
	\bibitem{saboori_2021_classification}
	R.~Saboori, A.~Salehi, P.~C. Vaidya, and K.~Dua, ``Classification of arrhythmia
	using machine learning techniques,'' pp.~445--452, 01 2021.
	
	\bibitem{hossain_2020_cardiac}
	M.~Hossain, S.~Of~Coe, M.~Tanseer, and A.~Of~Eee, ``Cardiac arrhythmia
	detection in an ecg beat signal using 1d convolution neural network 1 st,''
	2020.
	
	\bibitem{alrahhal_2018_convolutional}
	M.~M. Al~Rahhal, Y.~Bazi, M.~Al~Zuair, E.~Othman, and B.~BenJdira,
	``Convolutional neural networks for electrocardiogram classification,'' {\em
		Journal of Medical and Biological Engineering}, vol.~38, pp.~1014--1025, 03
	2018.
	
	\bibitem{sujadevi_2017_realtime}
	V.~G. Sujadevi, K.~P. Soman, and R.~Vinayakumar, ``Real-time detection of
	atrial fibrillation from short time single lead ecg traces using recurrent
	neural networks,'' {\em Advances in Intelligent Systems and Computing},
	pp.~212--221, 10 2017.
	
	\bibitem{faust_2018_automated}
	O.~Faust, A.~Shenfield, M.~Kareem, T.~R. San, H.~Fujita, and U.~R. Acharya,
	``Automated detection of atrial fibrillation using long short-term memory
	network with rr interval signals,'' {\em Computers in Biology and Medicine},
	vol.~102, pp.~327--335, 11 2018.
	
	\bibitem{wang_2019_a}
	G.~Wang, C.~Zhang, Y.~Liu, H.~Yang, D.~Fu, H.~Wang, and P.~Zhang, ``A global
	and updatable ecg beat classification system based on recurrent neural
	networks and active learning,'' {\em Information Sciences}, vol.~501,
	pp.~523--542, 10 2019.
	
	\bibitem{rajkumar_2019_arrhythmia}
	A.~Rajkumar, M.~Ganesan, and R.~Lavanya, ``Arrhythmia classification on ecg
	using deep learning,'' 03 2019.
	
	\bibitem{jo_2021_explainable}
	Y.-Y. Jo, Y.~Cho, S.~Y. Lee, J.-m. Kwon, K.-H. Kim, K.-H. Jeon, S.~Cho,
	J.~Park, and B.-H. Oh, ``Explainable artificial intelligence to detect atrial
	fibrillation using electrocardiogram,'' {\em International Journal of
		Cardiology}, vol.~328, pp.~104--110, 04 2021.
	
	\bibitem{sangroya_2022_generating}
	A.~Sangroya, S.~Jain, L.~Vig, C.~Anantaram, A.~Ukil, and S.~Khandelwal,
	``Generating conceptual explanations for dl based ecg classification model,''
	{\em The International FLAIRS Conference Proceedings}, vol.~35, 05 2022.
	
	\bibitem{bender_2023_analysis}
	T.~Bender, J.~M. Beinecke, D.~Krefting, C.~Müller, H.~Dathe, T.~Seidler,
	N.~Spicher, and A.-C. Hauschild, ``Analysis of a deep learning model for
	12-lead ecg classification reveals learned features similar to diagnostic
	criteria,'' {\em IEEE Journal of Biomedical and Health Informatics},
	pp.~1--12, 01 2023.
	
	\bibitem{gusarova_explainable}
	N.~Gusarova, I.~Tomilov, D.~Zmievskii, V.~Shilonosov, T.~Polevaya, and
	A.~Vatian, ``Explainable artificial intelligence in the diagnosis of
	cardiovascular diseases in small samples.''
	
	\bibitem{varandas_2022_quantified}
	R.~Varandas, B.~Gonçalves, H.~Gamboa, and P.~Vieira, ``Quantified
	explainability: Convolutional neural network focus assessment in arrhythmia
	detection,'' {\em BioMedInformatics}, vol.~2, pp.~124--138, 01 2022.
	
	\bibitem{neves_2021_interpretable}
	I.~Neves, D.~Folgado, S.~Santos, M.~Barandas, A.~Campagner, L.~Ronzio,
	F.~Cabitza, and H.~Gamboa, ``Interpretable heartbeat classification using
	local model-agnostic explanations on ecgs,'' {\em Computers in Biology and
		Medicine}, vol.~133, p.~104393, 06 2021.
	
	\bibitem{zhang_interpretable}
	D.~Zhang, S.~Yang, X.~Yuan, and P.~Zhang, ``Interpretable deep learning for
	automatic diagnosis of 12-lead electrocardiogram a b c,'' {\em iScience},
	Apr, 2021.
	
	\bibitem{pawar_2020_incorporating}
	U.~Pawar, S.~Rea, and R.~O\&apos;reilly, ``Incorporating explainable artificial
	intelligence (xai) to aid the understanding of machine learning in the
	healthcare domain,'' 2020.
	
	\bibitem{jo_2021_detection}
	Y.-Y. Jo, J.-m. Kwon, K.-H. Jeon, Y.-H. Cho, J.-H. Shin, Y.-J. Lee, M.-S. Jung,
	J.-H. Ban, K.-H. Kim, S.~Y. Lee, J.~Park, and B.-H. Oh, ``Detection and
	classification of arrhythmia using an explainable deep learning model,'' {\em
		Journal of Electrocardiology}, vol.~67, pp.~124--132, 07 2021.
	
	\bibitem{divakar_explainable}
	C.~Divakar, R.~Harsha, K.~Radha, V.~Dulipalla, Rao, N.~Madhavi, and
	T.~Bharadwaj, ``Explainable ai for cnn-lstm network in pcg-based valvular
	heart disease diagnosis explainable ai for cnn-lstm network in pcg-based
	valvular heart disease diagnosis.''
	
	\bibitem{le_2023_lightx3ecg}
	K.~H. Le, H.~H. Pham, T.~B. Nguyen, T.~A. Nguyen, T.~N. Thanh, and C.~D. Do,
	``Lightx3ecg: A lightweight and explainable deep learning system for 3-lead
	electrocardiogram classification,'' {\em Biomedical Signal Processing and
		Control}, vol.~85, p.~104963, 08 2023.
	
	\bibitem{Yol_2019_Design}
	Y.~Yol, M.~Özdemir, and A.~Akan, ``Design of real time cardiac arrhythmia
	detection device,'' 10 2019.
	
	\bibitem{goldberger_physiobank}
	A.~Goldberger, L.~Amaral, .~Glass, J.~Hausdorff, .~Plamen, C.~Ivanov, R.~Mark,
	J.~Mietus, G.~Moody, C.-K. Peng, and .~Stanley, ``Physiobank, physiotoolkit,
	and physionet components of a new research resource for complex physiologic
	signals.''
	
	\bibitem{ang_2023_a}
	G.~J.~N. Ang, A.~K. Goil, H.~Chan, X.~C. Lee, R.~B.~A. Mustaffa, T.~Jason,
	Z.~T. Woon, and B.~Shen, ``A novel application for real-time arrhythmia
	detection using yolov8,'' 05 2023.
	
	\bibitem{gao_2019_an}
	J.~Gao, H.~Zhang, P.~Lu, and Z.~Wang, ``An effective lstm recurrent network to
	detect arrhythmia on imbalanced ecg dataset,'' {\em Journal of Healthcare
		Engineering}, vol.~2019, pp.~1--10, 10 2019.
	
	\bibitem{singh_2018_sciencedirectncnd}
	S.~Singh, S.~Kumar~Pandey, U.~Pawar, and R.~Janghel, ``Sciencedirect-nc-nd
	license (https://creativecommons.org/licenses/by-nc-nd/3.0/) peer-review
	under responsibility of the scientific committee of the international
	conference on computational intelligence and data science (iccids 2018).
	classification of ecg arrhythmia using recurrent neural networks,'' {\em
		Procedia Computer Science}, vol.~00, pp.~0--000, 2018.
	
	\bibitem{raza_2022_designing}
	A.~Raza, K.~P. Tran, L.~Koehl, and S.~Li, ``Designing ecg monitoring healthcare
	system with federated transfer learning and explainable ai,'' {\em
		Knowledge-Based Systems}, vol.~236, p.~107763, 01 2022.
	
	\bibitem{howard_mobilenets}
	A.~Howard, M.~Zhu, B.~Chen, D.~Kalenichenko, W.~Wang, T.~Weyand, and
	M.~Andreetto, ``Mobilenets: Efficient convolutional neural networks for
	mobile vision applications.''
	
	\bibitem{sandler_mobilenetv2}
	M.~Sandler, A.~Howard, M.~Zhu, A.~Zhmoginov, and L.-C. Chen, ``Mobilenetv2:
	Inverted residuals and linear bottlenecks.''
	
	\bibitem{docherty_2003_12lead}
	B.~Docherty, ``12-lead ecg interpretation 2: right ventricular and posterior
	infarcts,'' {\em British Journal of Nursing}, vol.~12, pp.~1304--1311, 12
	2003.
	
	\bibitem{auricchio_2004_characterization}
	A.~Auricchio, C.~Fantoni, F.~Regoli, C.~Carbucicchio, A.~Goette, C.~Geller,
	M.~Kloss, and H.~Klein, ``Characterization of left ventricular activation in
	patients with heart failure and left bundle-branch block,'' {\em
		Circulation}, vol.~109, pp.~1133--1139, 03 2004.
	
	\bibitem{tzogias_2014_electrocardiographic}
	L.~Tzogias, L.~A. Steinberg, A.~J. Williams, K.~E. Morris, W.~J. Mahlow, R.~I.
	Fogel, J.~A. Olson, E.~N. Prystowsky, and B.~J. Padanilam,
	``Electrocardiographic features and prevalence of bilateral bundle-branch
	delay,'' {\em Circulation: Arrhythmia and Electrophysiology}, vol.~7,
	pp.~640--644, 08 2014.
	
\end{thebibliography}


\end{document}